\documentclass[letterpaper, 10 pt, conference]{ieeeconf}
\IEEEoverridecommandlockouts
\usepackage{comment}
\usepackage{amsmath}
\usepackage{graphicx}
\usepackage{hyperref}
\usepackage{float}
\usepackage[linesnumbered,ruled,vlined]{algorithm2e}
\usepackage{xcolor}
\usepackage{booktabs}
\usepackage{pifont}
\usepackage{amssymb}

\usepackage{caption}

\title{\LARGE \bf
Design and Control of Modular Magnetic Millirobots for Multimodal Locomotion and Shape Reconfiguration
}
 \author{Erik Garcia Oyono$^{1\dagger}$, Jialin Lin$^{1\dagger}$, and Dandan Zhang$^{1*}$% <-this % stops a space
\thanks{$^{1}$Department of Bioengineering, Imperial College London, London, United Kingdom. $\dagger$ These authors contributed equally to this work (co-first authors).*Corresponding author: {\tt\small d.zhang17@imperial.ac.uk}}%
}   

\begin{document}

\maketitle
\thispagestyle{empty}
\pagestyle{empty}

%%%%%%%%%%%%%%%%%%%%%%%%%%%%%%%%% ABSTRACT %%%%%%%%%%%%%%%%%%%%%%%%%%%%%%%%%
\begin{abstract}

% Problem/Context: Why modular robots are useful / what challenge exists
% Method/Approach: Description of robot design, control/planning algorithm
% Results: Experimental demonstrations, validation/ observations
% Implications: Broader relevance (scalability, manufacturing, medical)

%Modular small-scale robots can assemble and disassemble on demand, enabling task-specific adaptation in dynamic environments. However, most existing platforms rely on workspace collisions for reconfiguration, bulky electromagnetic setups, and neglect single-module control, limiting their translation to real-life settings.

Modular small-scale robots offer the potential for on-demand assembly and disassembly, enabling task-specific adaptation in dynamic and constrained environments. However, existing modular magnetic platforms often depend on workspace collisions for reconfiguration, employ bulky three-dimensional electromagnetic systems, and lack robust single-module control, which limits their applicability in biomedical settings.
In this work, we present a modular magnetic millirobotic platform comprising three cube-shaped modules with embedded permanent magnets, each designed for a distinct functional role: a \textit{free} module that supports self-assembly and reconfiguration, a \textit{fixed} module that enables flip-and-walk locomotion, and a \textit{gripper} module for cargo manipulation. Locomotion and reconfiguration are actuated by programmable combinations of time-varying two-dimensional uniform and gradient magnetic field inputs.
Experiments demonstrate closed-loop navigation using real-time vision feedback and A* path planning, establishing robust single-module control capabilities. Beyond locomotion, the system achieves self-assembly, multimodal transformations, and disassembly at low field strengths. Chain-to-gripper transformations succeeded in 90\% of trials, while chain-to-square transformations were less consistent, underscoring the role of module geometry in reconfiguration reliability.
These results establish a versatile modular robotic platform capable of multimodal behavior and robust control, suggesting a promising pathway toward scalable and adaptive task execution in confined environments.

\end{abstract}

%%%%%%%%%%%%%%%%%%%%%%%%%%%%%%%%%%% INTRO %%%%%%%%%%%%%%%%%%%%%%%%%%%%%%%%%%
\section{Introduction}

\begin{figure}
    \centering
    \vspace{-0.1cm}
\captionsetup{font=footnotesize,labelsep=period}
    \includegraphics[width=\linewidth]{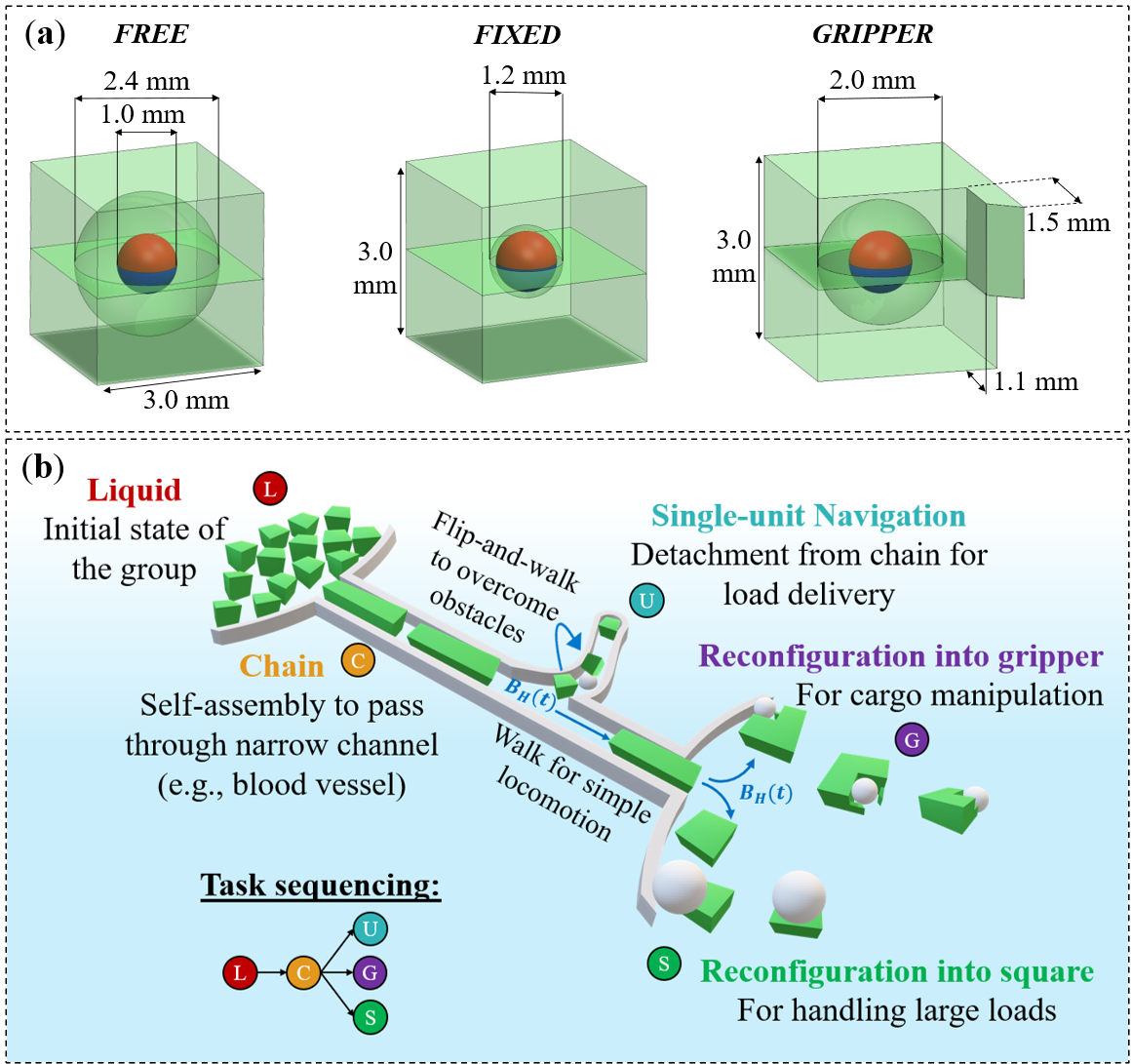}
 %   \caption{Conceptual overview of the modular magnetic millirobot platform. (a) Design and dimensions of the three cube-shaped modules (\textit{free}, \textit{fixed}, and \textit{gripper}) with embedded permanent magnets. (b) Illustration of multimodal capabilities enabled by magnetic actuation: modules can navigate individually, self-assemble into chains, reconfigure for cargo manipulation, and disassemble on demand.}
 \caption{Conceptual overview of the modular magnetic millirobot platform. (a) Geometry and key dimensions of the three cube-shaped modules (\textit{free}, \textit{fixed}, and \textit{gripper}), each incorporating an embedded permanent magnet with module-specific orientation. (b) Multimodal, field-programmable behaviors enabled by magnetic actuation: individual locomotion and navigation, self-assembly into chains, reconfiguration for cargo grasping/manipulation, and on-demand disassembly.}
    \label{fig:Modules}
    \vspace{-0.4cm}
\end{figure}

In recent decades, small-scale robots have emerged as promising tools for biomedical applications, enabling minimally invasive access and precise interaction within confined anatomical spaces \cite{Lee2023,zhang2023advanced}. Their capabilities have been explored across diverse tasks, including in vivo diagnostics using capsule-like platforms \cite{castellanos2024multimodal,yu2025hybrid}, site-specific drug delivery \cite{wang2022microrobots,liu2022untethered}, cell and tissue manipulation \cite{Wang2024,zhang2022fabrication}, and microsurgical procedures that demand controlled navigation and manipulation at small scales \cite{jiang2024digital}. Despite these advances, most existing systems are engineered for a single predefined function and fixed morphology. While effective within their respective domains, such task-specific designs often lack the adaptability required in complex physiological environments such as vascular networks, where a single mission may require sequential capabilities including confined-space navigation, cargo transport, and targeted manipulation.

%In recent decades, small-scale robots have emerged as promising tools for biomedical applications, offering minimally invasive access and multifunctional task execution \cite{Lee2023,zhang2023advanced}. Traditional designs, such as capsule-like microrobots for diagnostics \cite{castellanos2024multimodal}, microrobots for site-specific drug delivery \cite{wang2022microrobots,liu2022untethered}, and gripping microrobots for cell manipulation \cite{Wang2024,zhang2022fabrication}, are often optimized to perform a single function. While effective within their respective domains, these devices lack the adaptability required in complex environments such as vascular networks, where sequential tasks, including confined-space navigation, cargo transport, and targeted manipulation, are essential.

To address this limitation, the concept of transformable modular microrobots has gained increasing attention \cite{zhang2024sonotransformers,xie2019reconfigurable}. Modular robotic systems are composed of multiple interconnected units capable of dynamic reconfiguration, enabling structural transformation in response to environmental or task demands \cite{Ahmadzadeh2015}. This modular architecture supports \textit{multimodal transformations}, allowing a single robotic system to switch between different morphologies and functionalities, for example, transitioning from a streamlined configuration for navigation to an expanded or articulated structure for manipulation. Such transformability provides a pathway toward enhanced adaptability and multifunctionality at the microscale.
However, translating modularity into biomedical microrobotics introduces substantial technical challenges. At micro- and nano-scales, the integration of onboard sensors, power sources, and actuators is severely constrained by fabrication limits and size restrictions \cite{Fiaz2019}. 

To address these constraints, researchers have explored various external wireless actuation strategies, including optical \cite{zhang2020distributed}, magnetic \cite{lin2024magnetic}, acoustic \cite{deng2023acoustically}, electric \cite{zhuang2025optoelectronic}, and hybrid field approaches \cite{wang2025advanced}. Among these, magnetic actuation is particularly well suited for transformable modular microrobots. Magnetic fields provide deep tissue penetration, efficient wireless power transmission, and the capability to simultaneously control multiple modules without physical tethering \cite{Jiang2022}. These advantages make magnetic actuation an effective foundation for coordinated assembly, disassembly, and reconfiguration in complex in vivo environments.

\begin{table*}[t]
\centering
\caption{Comparison with representative modular magnetic systems.}
\resizebox{\textwidth}{!}{
\begin{tabular}{lcccccccc}
\toprule
\textbf{System} & \textbf{Actuation} & \textbf{Max Field (mT)} & \textbf{Module ($mm^3$)} & \textbf{Collision-free Reconfig.} & \textbf{Single-module Closed-Loop Control} & \textbf{Workspace ($mm^2$)} & \textbf{Cooling Req.}\\
\midrule
Rogowski et al. 2020 \cite{Rogowski2020} & 3D & 40 & $26 \times 4 \times 4$ & -- & \ding{55} & $100 \times 100$ & \checkmark \\
Bhattacharjee et al. 2022 \cite{Bhattacharjee2022} & 3D & 22 & $2.8 \times 2.8 \times 2.8$ & \ding{55} & \ding{55} & $110 \times 110$ & \ding{55} \\
Lu et al. 2023 \cite{Lu2023} & 3D & 23 & $2.8 \times 2.8 \times 2.8$ & \ding{55} & \ding{55} & $33 \times 33$ & \ding{55} \\
Rogowski et al. 2023 \cite{Rogowski2023} & 3D & 32 & $4 \times 4 \times 4$ & -- & \ding{55} & $100 \times 100$ & \checkmark \\
\textbf{This work} & \textbf{2D} & \textbf{13} & \boldmath{$3 \times 3 \times 3$}
 & \checkmark & \checkmark & \textbf{$35 \times 35$} & \ding{55} \\
\bottomrule
\end{tabular}}
\label{tab:SOTA}
\end{table*}

Within the domain of magnetic modular microrobots, researchers have progressively pushed modularity toward enhanced adaptability under globally applied fields. Rogowski \emph{et al.} introduced programmable magnetic cuboid modules with embedded magnets that assemble and collaborate to transport cargo, illustrating how simple magnetic architecture can yield coordinated behaviors \cite{Rogowski2020}. Building on this, Bhattacharjee \emph{et al.} demonstrated open-loop shape-programmed assemblies of magnetic modules, where face magnetization and field sequences produced predictable assembled configurations under uniform fields. Closed-loop approaches were subsequently explored by Lu \emph{et al.}, who coupled visual feedback with field control to improve assembly repeatability and robustness \cite{Lu2023}. Rogowski \emph{et al.} achieved closed-loop navigation of modular chains, advancing controllability of reconfigurable systems and illustrating that modular assemblies can be steered in a planned manner under visual feedback and field modulation.

Although these efforts represent meaningful progress toward controllable self-assembling and reconfigurable magnetic systems, several challenges remain. First, many reconfiguration strategies rely on boundary interactions or collisions to induce morphology changes, which can be undesirable or unsafe in sensitive environments such as biological tissue. Second, achieving independent control of individual modules within an assembly is difficult when all units are driven by the same global magnetic field. In these cases, differentiated motion is typically obtained by exploiting geometric or magnetization differences between robots rather than through true addressable control, where each unit can be actuated independently under a shared input \cite{Razzaghi2023}. Third, the actuation infrastructure commonly used in modular magnetic robotics involves complex multi-axis coil systems that must generate relatively high magnetic fields (10–40 mT) to produce usable torques and forces at millimeter scales, resulting in high power consumption, significant resistive heating, and bulky hardware \cite{Yang2020}.

%Taken together, these limitations emphasize the need to reduce the required magnetic field strength while still achieving precise multimodal control, including safe reconfiguration and the ability to actuate individual modules, as an important step toward practical biomedical magnetic robotic platforms.

In this work, we present a modular magnetic millirobotic platform that integrates self-reconfiguration, multimodal transformations, and single-unit control under a low-field electromagnetic setup (\autoref{fig:Modules}). Unlike prior approaches that depend on boundary interactions for reconfiguration, our system enables controlled transformations through programmed magnetic field sequences, improving operational safety in constrained environments.

The contributions of this work are as follows:
\begin{itemize}
    \item \textbf{Develop a modular millirobotic system} consisting of three complementary units (\textit{free}, \textit{fixed}, and \textit{gripper}) together with a hardware-efficient 2D coil setup, enabling both single-unit locomotion and cooperative assembly.
    \item \textbf{Demonstrate locomotion and multimodal transformations} under low magnetic fields ($<13$ mT), eliminating reliance on boundary-dependent reconfiguration.
    \item \textbf{Achieve closed-loop single-module control} using vision feedback and path planning, enabling adaptive navigation in constrained environments.
\end{itemize}

Together, these advances mitigate major limitations of existing modular magnetic systems and provide a foundation for developing versatile, scalable, and translatable millirobotic platforms for real-world applications.

%%%%%%%%%%%%%%%%%%%%%%%%%%%%%% System Design %%%%%%%%%%%%%%%%%%%%%%%%%%%%%%
\section{System Design}

\subsection{Actuation Principles}

\begin{figure*}[t]
    \centering
    \captionsetup{font=footnotesize,labelsep=period}
        \vspace{-0.1cm}
\captionsetup{font=footnotesize,labelsep=period}
    \includegraphics[width=\textwidth]{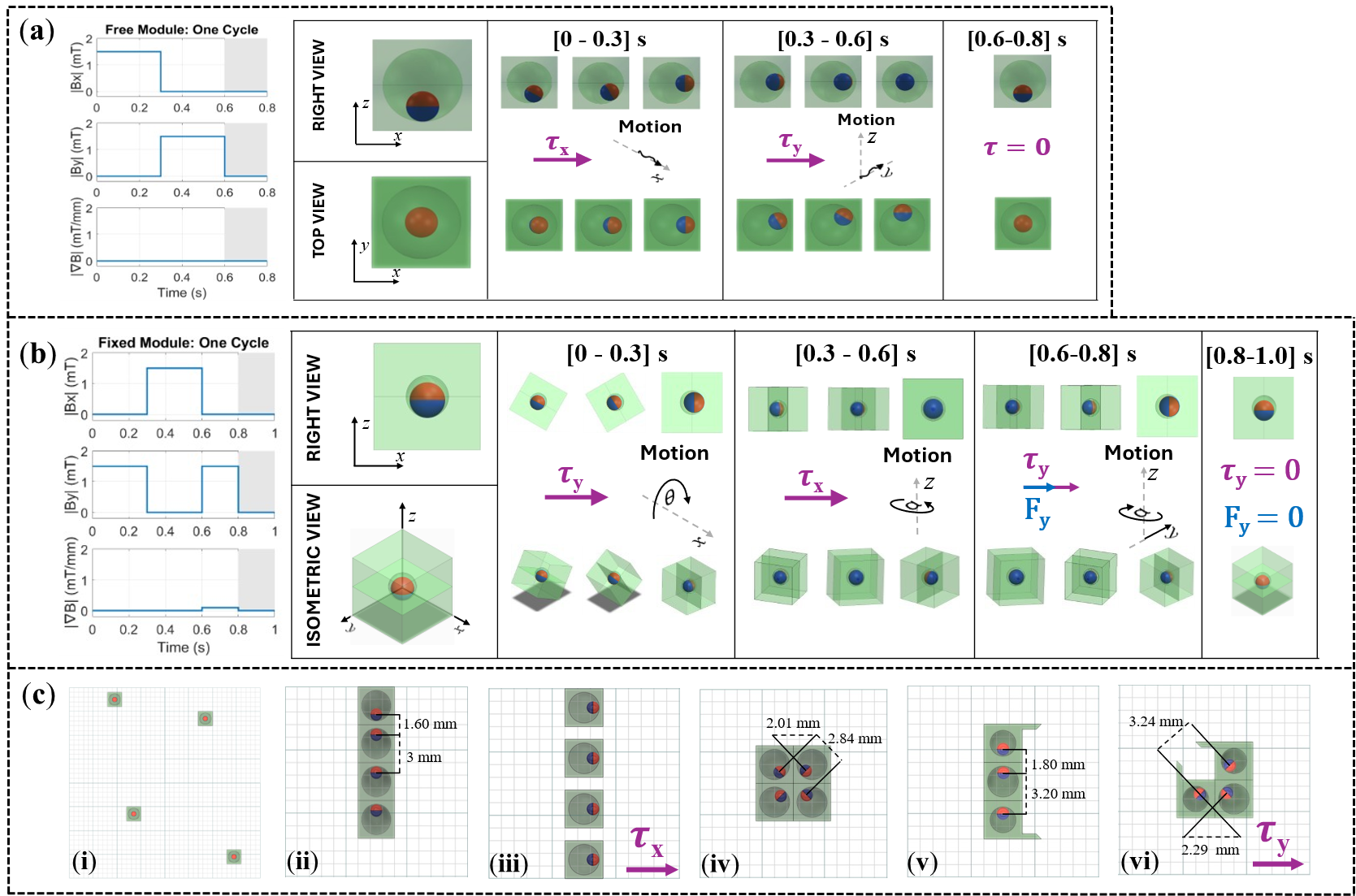}
    \caption{Actuation mechanisms. (a) Single-module locomotion of the \textit{free} module driven by uniform fields. (b) Single-module locomotion of the \textit{fixed} module driven by combined uniform and gradient fields. (c) Cuboid modules exhibit field-programmable self-organization: (i) In the absence of an external field, the dipoles align along the $z$-axis and the modules remain separated (``liquid'' state). (ii) When brought into close proximity ($<7$~mm), dipole-dipole attraction dominates and modules self-assemble into chains. (iii) Superimposing a strong external field perpendicular to the chain direction aligns the dipoles with the applied field, suppressing chaining and maintaining separation. (iv) Appropriate field modulation yields stable ``square'' assemblies. (v) A chain composed of two \textit{gripper} modules and one \textit{free} module forms through dipole–dipole attraction, creating a linear three-module assembly.
(vi) Upon application of a sufficiently strong and rapid uniform-field pulse, the weakest dipole bond in the chain is selectively broken, enabling perpendicular realignment and reattachment of the modules into a functional gripper configuration.}
\label{fig:Motion}
    % \caption{Actuation Mechanisms. (a) Single-module motion of the \textit{free} module, using only uniform fields. (b) Single-module motion of the \textit{fixed} module, using uniform and gradient fields. (c) The cuboid robots can organize into different patterns depending on their spacing and the external MF applied: (i) Under no external field, the dipoles are aligned along the $z$-axis and remain separate, representing their “liquid” state. (ii) In close proximity ($<7 mm$), the modules self-assemble into a “chain” due to dominant dipolar attraction, but when a strong perpendicular external MF is superimposed (iii), the dipoles remain aligned with the applied field, preventing chaining and maintaining separation. (iv) Controlled field modulation enables formation of stable “square” assemblies. In addition, chains comprising two gripper modules and one free module (v) can be reconfigured into functional grippers (vi) when a sufficiently strong and rapid uniform field is applied to overcome the weakest dipole bond, enabling perpendicular reattachment.}    \label{fig:Motion}
      \vspace{-0.4cm}
\end{figure*}

Controlled locomotion and reconfiguration of the millirobotic modules are achieved through the coordinated use of uniform (Helmholtz) magnetic fields and gradient (Maxwell) magnetic fields. Uniform fields primarily generate magnetic torques that align the magnetic dipole moments of the modules with the applied field direction, enabling orientation control and rotational motion. In contrast, gradient fields produce net magnetic forces that drive translational motion by pulling the modules along the field gradient \cite{Wang2023}.

Beyond field-induced torque and force, magnetic dipole-dipole interactions between neighboring modules play a crucial role in system-level behavior. These interactions enable reversible self-assembly and controlled disassembly, allowing modules to form chains or detach in response to programmed field sequences \cite{Yang2021}.

\subsubsection{Single-Module Actuation}
A magnetic dipole $\mathbf{m}$ placed in a uniform field $\mathbf{B}_H$ experiences a torque
\begin{equation}
    \boldsymbol{\tau} = \mathbf{m} \times \mathbf{B}_H,
    \label{eq:Torque}
\end{equation}
which tends to align the dipole with the field direction. 
In a gradient field $\nabla \mathbf{B}_M$, it additionally experiences a force
\begin{equation}
    \mathbf{F} = (\mathbf{m} \cdot \nabla)\mathbf{B}_M,
\end{equation}
producing translation.

Intuitively, the magnetization layout and mechanical constraints of each module determine its dominant actuation mode:

\begin{itemize}
\item \textbf{\textit{Free} module: stick-slip translation.} The magnet is loosely housed and can oscillate within an internal cavity. Under an oscillatory uniform-field torque, the magnet periodically contacts the cavity walls and the substrate. This cyclic torque is rectified through frictional interactions, producing net stick-slip displacement of the module (\autoref{fig:Motion}(a)).
\item \textbf{\textit{Fixed} module: flip-and-walk locomotion.} The magnet is rigidly embedded, so the applied torque rotates the entire cube rather than the magnet alone. Alternating torque-driven flipping and gradient-induced biasing break symmetry and yield directional “flip-and-walk” motion (\autoref{fig:Motion}(b)).
\item \textbf{\textit{Gripper} module: reconfiguration-oriented docking.} This module is not intended for locomotion. Instead, its magnet orientation is selected to favor stable, perpendicular dipole alignment during assembly, promoting reliable docking and reconfiguration (\autoref{fig:Motion}(c)).
\end{itemize}

\subsubsection{Multi-Module Interactions}
When multiple dipoles are present, each produces fields that affect its neighbors. For two dipoles $\mathbf{m}_1$ and $\mathbf{m}_2$ separated by vector $\mathbf{r}$, the induced interaction force can be decomposed into radial and tangential components:
\begin{align}
    F_r &= \frac{3\mu_0 m_1 m_2}{4\pi r^4}(1 - 3\cos^2\theta), \\
    F_\theta &= \frac{3\mu_0 m_1 m_2}{4\pi r^4}(2\cos\theta\sin\theta)
\end{align}
where $\theta$ is the angle between the dipole axis and $\mathbf{r}$. 

This decomposition highlights the well-known “magic angle” at $\theta=54.7^\circ$: below this angle, radial attraction dominates, driving chain assembly; above it, repulsion dominates, enabling separation \cite{Yang2021}. By reorienting dipoles with external fields, one can program modules to self-assemble into chains, reconfigure into new geometries, or disassemble on demand (\autoref{fig:Motion}(c)).

% 2.1 EMS
\subsection{Hardware Design}
\subsubsection{Electromagnetic System}

% [Q] Similar papers add an image of the reference coordinate system for EMS/modular cube, should I add this? I already made one for the FSM angle update logic, but don't really talk about it
\begin{figure}
    \centering
\captionsetup{font=footnotesize,labelsep=period}
    \includegraphics[width=\linewidth]{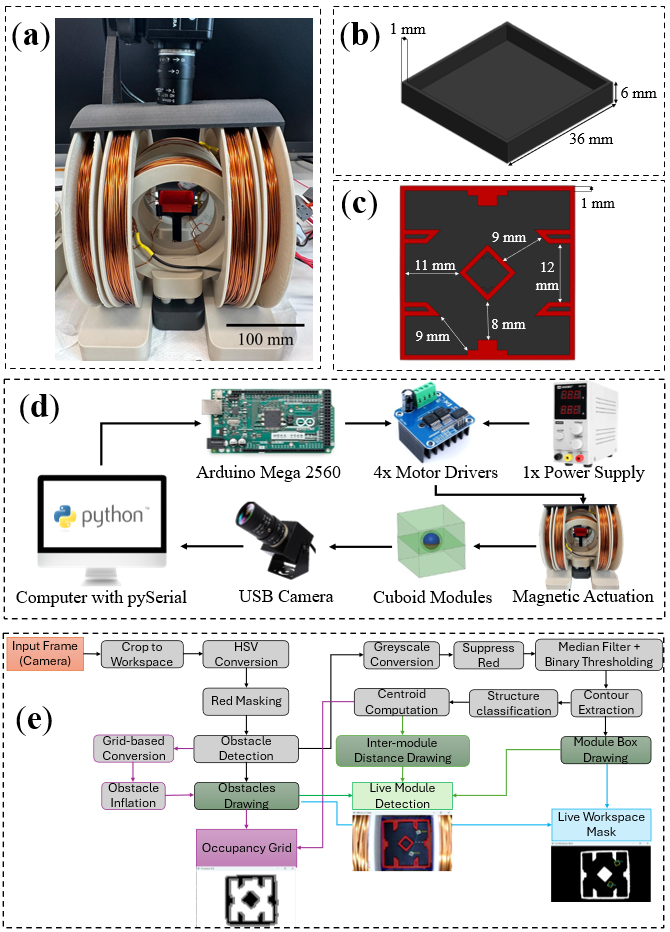}
 %   \caption{(a) Two-dimensional Helmholtz and Maxwell electromagnetic setup. (b) Planar workspace. (c) Maze navigation workspace. (d) Experimental setup. (e) Real-time detection pipeline and live window outputs. The system supports live processing and recording from a camera feed, single-image analysis, and batch processing of datasets.}
 \caption{Experimental and control setup. (a) Two-dimensional electromagnetic actuation system combining Helmholtz (uniform) and Maxwell (gradient) coil pairs. (b) Planar workspace used for locomotion and assembly experiments. (c) Maze workspace for constrained navigation trials. (d) Photograph of the full experimental apparatus. (e) Real-time vision pipeline and live output windows; the software supports live camera processing and recording, single-image inference, and batch analysis of datasets.}
    \label{fig:EMS}
      \vspace{-0.1cm}
\end{figure}

A two-dimensional electromagnetic system comprising orthogonal Helmholtz and Maxwell coil pairs along the $X$ and $Y$ axes generates uniform fields and linear gradients (\autoref{fig:EMS}(a)). The magnetic field distribution can be expressed as:

\begin{equation}
\label{eq:EMS}
\mathbf{B}(\mathbf{r}) = 
\begin{bmatrix}B_x \\ B_y \\ 0\end{bmatrix} +
\begin{bmatrix}G_x-\frac{G_y}{2} & 0 & 0 \\ 0 & -\frac{G_x}{2}+G_y & 0 \\ 0 & 0 & -\frac{G_x+G_y}{2}\end{bmatrix}\mathbf{r}
\end{equation}
where $B_{x,y} = k_{Hx,Hy} I_{Hx,Hy}$ and $G_{x,y} = k_{Mx,My} I_{Mx,My}$ are linearly mapped from coil currents using experimentally calibrated constants. The system satisfies $\nabla \cdot \mathbf{B} = 0$, and the field is approximately 2D near $z\approx 0$, allowing planar experiments.

Coil frames, camera mount, and workspace components were 3D-printed in PLA using a Bambu Lab P1S printer with a 0.4 mm nozzle. Each coil was fabricated by winding 1 mm enameled copper wire around 14 mm thick 3D-printed PLA frames. Helmholtz coils were separated by their radius and driven with co-directional currents, while Maxwell coils were spaced by $\sqrt{3}\times radius$ radius and operate with opposing currents. Physical specifications are summarized in Table~\ref{tab:coil_specs}.

\begin{table}
\captionsetup{font=footnotesize,labelsep=period}
\caption{Physical specifications of the Helmholtz and Maxwell pairs along the $x$- and $y$-axis, respectively.}
\centering
\begin{tabular}{lcccc}
\toprule
\textbf{Coil} & \textbf{Diameter (m)} & \textbf{Spacing (m)} & \textbf{Turns} \\
\midrule
Hx & 0.100 & 0.050 & 100 \\
Hy & 0.176 & 0.088 & 176 \\
Mx & 0.104 & 0.090 & 100 \\
My & 0.152 & 0.132 & 152 \\
\bottomrule
\end{tabular}
\label{tab:coil_specs}
  \vspace{-0.5cm}
\end{table}

The workspace, located at the center of the electromagnetic system, measured $35\times35$~mm$^2$ and was 3D-printed in black to enable contrast-based module detection (\autoref{fig:EMS}(b)). A second maze-shaped workspace was manufactured with red obstacles for boundary detection (\autoref{fig:EMS}(c)).

The experimental setup is depicted in \autoref{fig:EMS}(d). The system was powered by an LW-K3010D 24 V programmable supply. An Arduino MEGA 2560 generated PWM signals to four BTS7960 motor drivers. Each motor driver regulates the current in one pair of coils. A USB camera mounted on a 3D-printed frame captured the workspace at 30 fps, and a white LED strip arranged in a square around the camera lens provides uniform illumination.

% 2.2 Robots
\subsubsection{Design and Fabrication of Cuboid Millirobots}

Three module types (\textit{free}, \textit{fixed}, \textit{gripper}) were designed in SolidWorks with base dimensions $3 \times 3 \times 3$ $mm^3$ and embedded 1 mm spherical N40 magnets (\autoref{fig:Modules}(a)). \textit{Free} modules have a 2.4 mm cavity allowing magnet rotation; \textit{fixed} modules constrain internal magnet motion (1.2 mm cavity) for deterministic locomotion; \textit{gripper} modules have a 2.0 mm cavity and a fin for grasping when assembled.

The modules were SLA-printed in clear resin (Formlabs 3B, 50 \( \mu m\) resolution), washed in isopropyl alcohol, and assembled using a custom 3D-printed mold with precise alignment. Micro-precise Gorilla Superglue fixed the halves, dried for 8 hours, and modules were spray-painted matte white for vision tracking.
% [Q] Should I add image of manufacturing method in the appendix?

\subsection{Software Design}
\subsubsection{Real-Time Detection}

A computer vision-based pipeline detects and tracks modules and obstacles within the workspace (\autoref{fig:EMS}(e)). Red obstacles are masked using HSV thresholding and inflated to generate a binary occupancy grid for path planning. Modules are identified from grayscale images using contour area and aspect ratio classification, and trajectories, initial separation, and reconfiguration events are recorded for analysis.

% 4.3 Closed-Loop Maze Navigation
\subsubsection{Closed-Loop Maze Navigation}

Closed-loop control of the \textit{fixed} module was implemented to navigate a maze environment using real-time vision feedback and path planning. We represent the workspace as an occupancy grid and employ an 8-connected A* search algorithm to generate collision-free paths. The A* algorithm is widely used in robotics for grid-based navigation because it efficiently balances accumulated travel cost and a heuristic estimate of remaining distance, producing optimal or near-optimal paths in static, known environments such as the constrained maze used in this work \cite{lim2020analysis}.

Planned cardinal and diagonal moves from A* are mapped to coil current commands through a finite state machine (FSM) that accounts for the module’s current orientation. After each motion command, the system monitors actual progress; if the module stalls or fails to follow the expected trajectory, the controller incrementally increases the applied field strength in 0.1 A steps to overcome local resistance and maintain forward motion (\autoref{alg:fsm}). 

\begin{algorithm}
\DontPrintSemicolon
\SetAlgoLined
\KwIn{Path $\mathcal{P} = \{p_1, \dots, p_n\}$, orientation $\mathbf{O}$}
\KwOut{Module reaches $p_n$}

\For{$p \in \mathcal{P}$}{
    $\mathbf{u} \gets$ ComputeDirection(current position, $p$)\;
    
    \If{first step}{ ApplyHelmholtzPulse($\mathbf{u}$) }
    \Else{ ApplyHelmholtzMaxwell($\mathbf{u}$) }

    $r \gets 0$\;
    \While{Distance(current, $p$) $>$ tolerance \textbf{and} $r < r_\text{max}$}{
        IncreaseField($\Delta I$)\;
        Reapply($\mathbf{u}$)\;
        $r \gets r+1$\;
    }

    \If{Distance(current, $p$) $\leq$ tolerance}{ UpdateOrientation($\mathbf{O}, \mathbf{u}$) }
    \Else{ AbortMotion(); \textbf{break} }
}

\caption{\textit{FSM-Controlled Navigation Loop}}
\label{alg:fsm}

\end{algorithm}

%%%%%%%%%%%%%%%%%%%%%%%%%%%%%%%%%%%% RESULTS %%%%%%%%%%%%%%%%%%%%%%%%%%%%%%%%%%%
\section{Experiments and Results}

This section presents experimental validation of the proposed modular magnetic millirobotic platform. We first characterize single-module locomotion of the \textit{free} and \textit{fixed} modules to evaluate the effect of uniform and gradient fields on translational motion. Next, we demonstrate multimodal transformations including self-assembly, chain-to-gripper and chain-to-square reconfiguration, and disassembly under low-field actuation. Finally, we test closed-loop navigation of a single \textit{fixed} module in a maze using real-time vision feedback and A* path planning.

\begin{figure}
    \centering
    \captionsetup{font=footnotesize,labelsep=period}
    \includegraphics[width=1\linewidth]{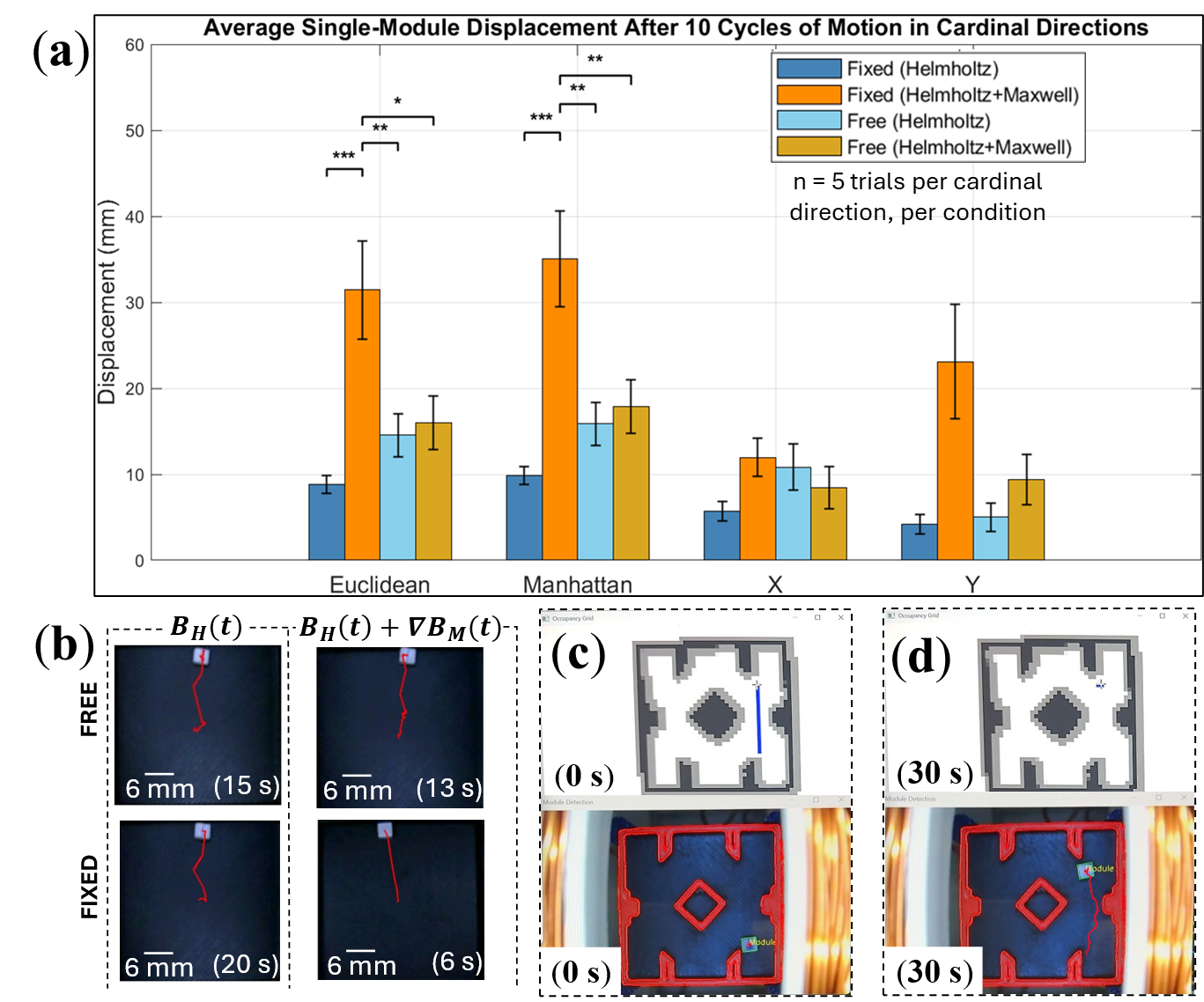}
%    \caption{Single-module translation of custom modules. (a) Mean displacement metrics (Euclidean, Manhattan, X, and Y) and their standard deviation after 10 cycles, recorded over five trials per cardinal direction (80 trials total across four directions and four conditions). (b) Trajectories generated by the (top) \textit{free} and (bottom) \textit{fixed} modules in response to an “UP” command in the presence of uniform fields only (left) and uniform fields superimposed with gradient fields (right). Both modules moved in the intended directions under both conditions, but the presence of gradient fields generated straight-line trajectories more consistently. (c) Live outputs of the occupancy grid (top) and live workspace mask (bottom) during goal setting. (d) Trajectory generated by the \textit{fixed} module through closed-loop control.}
\caption{Single-module translation and closed-loop navigation. (a) Mean displacement per direction (Euclidean, Manhattan, $x$, and $y$) with standard deviation after 10 actuation cycles, averaged over five trials per cardinal direction (80 trials total across four directions and four field conditions). (b) Representative trajectories for an ``UP'' command produced by the \textit{free} (top) and \textit{fixed} (bottom) modules under uniform fields only (left) and uniform fields with superimposed gradient fields (right). Both modules move in the commanded direction in all cases, while added gradient fields yield more repeatable straight-line motion. (c) Live goal-setting interface showing the occupancy grid (top) and workspace mask (bottom). (d) Closed-loop trajectory of the \textit{fixed} module during autonomous navigation.}
    \label{fig:Actuation}
          \vspace{-0.4cm}
\end{figure}

% 5.1 Single-Module Actuation
\subsection{Single-Module Locomotion}

Independent actuation was achieved with uniform fields of 1.5 mT. Unlike previous reports that required chains for meaningful locomotion \cite{Rogowski2023}, these results demonstrate controllable single-unit translation via time-varying uniform field modulation and slip-mediated interactions, which can be advantageous for confined-space navigation or targeted cargo delivery.

\autoref{fig:Actuation}(a) quantifies module displacement after ten commanded cycles. Four metrics are reported: (i) Euclidean displacement, which measures the straight-line distance between start and end positions; (ii) Manhattan displacement, which sums the absolute distances traveled in $x$ and $y$ directions and therefore captures non-linear paths; (iii) $x$ displacement, which isolates motion along the commanded axis; and (iv) $y$ displacement, which measures displacement orthogonal to the commanded axis. Together, these metrics allow us to assess both locomotion efficiency (Euclidean vs. Manhattan) and directional accuracy (X vs. Y).

The graph shows that the \textit{fixed} module translates reproducibly and is strongly affected by the addition of gradient fields, whereas the \textit{free} module exhibits more erratic motion and no statistically different change with gradients. Asterisks indicate statistical differences between \textit{H} and \textit{H+M} conditions (Tukey's Honestly Significant Difference test, $\alpha = 0.05$).

The \textit{free} module exhibited repeatable but more erratic translations in four directions, which could be partially corrected using gradient fields. The \textit{fixed} module achieved deterministic “flip-and-walk” trajectories in eight directions when gradient fields were applied (e.g., \autoref{fig:Actuation}(b)). After 10 cycles, Euclidean displacement of the \textit{fixed} module increased from $8.84 \pm 1.02$ mm under Helmholtz fields to $31.44 \pm 5.73$ mm with Helmholtz+Maxwell ($p=0.00013$). By contrast, gradients did not statistically alter displacement of the free module ($p>0.05$).

Side-by-side comparison shows that cavity geometry determined whether a module favored reconfiguration (\textit{free}) or deterministic translation (\textit{fixed}). Thus, cavity design can be tuned to prioritize either flexibility for reconfiguration or stability for directed locomotion, complementing recent work on structural optimization of modular microrobots \cite{Jiang2022}.

\begin{figure}
    \centering
        \captionsetup{font=footnotesize,labelsep=period}
    \includegraphics[width=\linewidth]{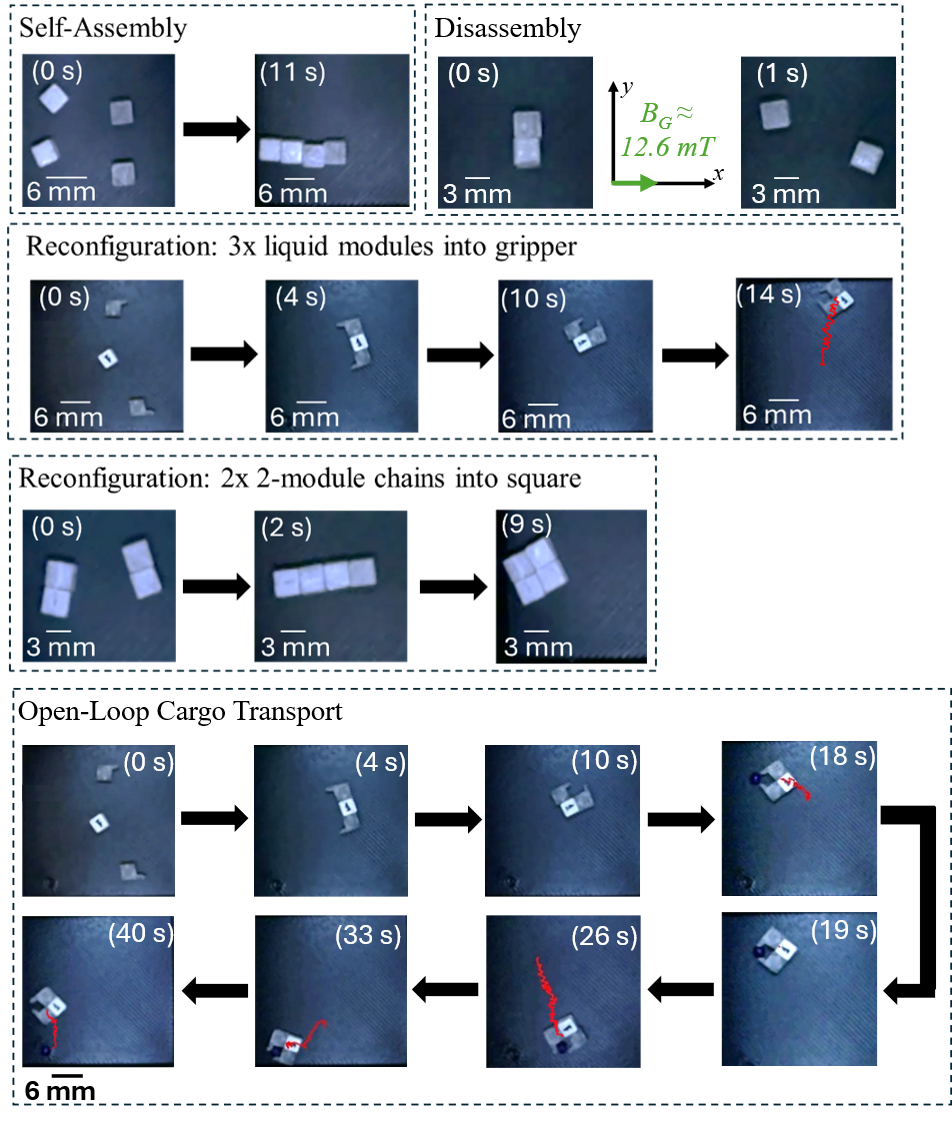}
    \caption{Field-programmable self-assembly, reconfiguration, and cargo transport. 
Top row: (Self-assembly) Discrete modules transition from a separated ``liquid'' state to a chain configuration under controlled magnetic fields. (Disassembly) Reversal of the field conditions separates the assembled chain into individual modules. 
Second row: (Reconfiguration: 3 liquid modules into gripper) Three initially separated modules are sequentially assembled and reoriented to form a functional gripper through programmed field modulation. 
Third row: (Reconfiguration: two 2-module chains into square) Two independent chains merge and rearrange into a stable square assembly. 
Bottom row: (Open-loop cargo transport) A reconfigured assembly manipulates and transports cargo along a prescribed path under open-loop magnetic control. Timestamps indicate elapsed time, and scale bars denote 3~mm or 6~mm as labeled.}
   % \caption{Multimodal transformations enabled by the system: Self-assembly into chain, disassembly of modular structures, chain-to-gripper reconfiguration, chain-to-square reconfiguration, disassembly, and cargo manipulation open-loop sequence.}
    \label{fig:MMT}
      \vspace{-0.4cm}
\end{figure}

% 5.2 Multimodal Transformations
\subsection{Multimodal Transformations}

\begin{figure} % WIP: Editing in PowerPoint
    \centering
        \captionsetup{font=footnotesize,labelsep=period}
    \includegraphics[width=\linewidth]{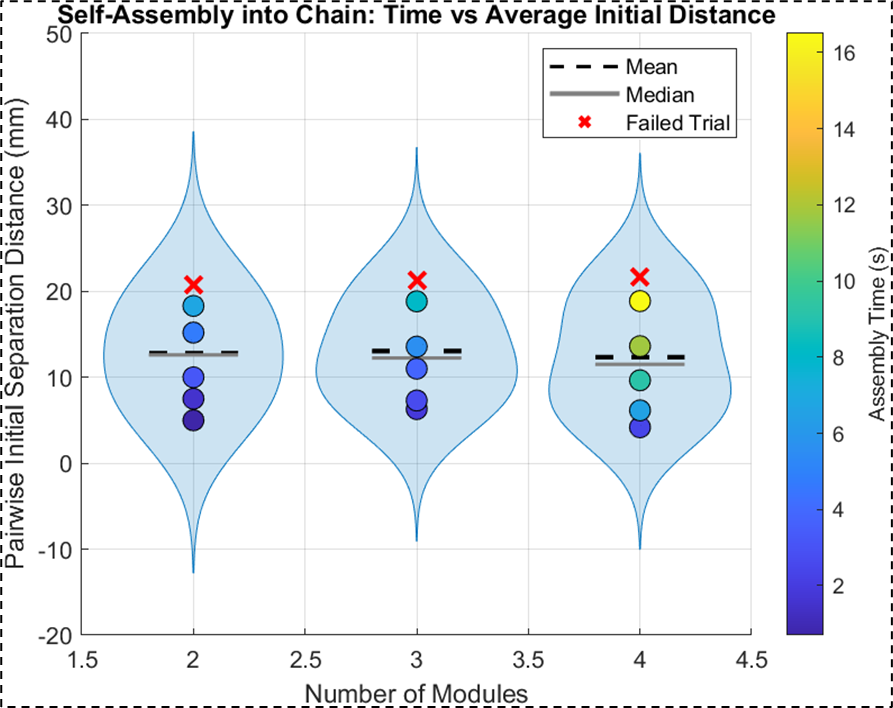}
   % \caption{Plot comparison of the initial separation distance against time for self-assembly of \textit{free} modules into a chain, where the blue shaded area represents the distribution of pairwise initial separation distances for each group of modules. As indicated by the color scale, the assembly time increases with both the number of modules and the initial separation distance, with failed trials (red crosses) occurring more frequently at larger distances. Dashed black lines denote the mean, while solid gray lines denote the median values for each group.}
   \caption{Self-assembly time versus initial separation distance for \textit{free} modules forming chains. The blue shaded region shows the distribution of pairwise initial separation distances within each module group. Consistent with the color scale, assembly time increases with both the number of modules and the initial separation distance; failed trials (red crosses) occur predominantly at larger separations. Dashed black lines indicate the mean, and solid gray lines indicate the median for each group.}
    \label{fig:ChTID}
      \vspace{-0.4cm}
\end{figure}

The multimodalities of the system presented in this study are depicted in \autoref{fig:MMT}. Compared to existing systems that rely on boundary collisions to trigger reconfiguration (\autoref{tab:SOTA}), this work achieved shape transformations without contacting workspace walls. This application is beneficial for biomedical translation, where collision-based reconfiguration could damage delicate tissues. The ability to toggle dipole interactions by modulating weak fields suggests a promising approach to safe, controllable transformations \textit{in vivo}.

\subsubsection{Self-Assembly into Chains}
% Self-assembly into chains occurred at 1.5 mT, with assembly time scaling approximately linearly with initial separation distance (\autoref{fig:MMT}(a)). Chains, grippers, and squares were reconfigured under low fields of 1.5-2.5 mT, while disassembly occurred at $\approx 12.6$ mT. These field strengths are substantially lower than the 10-40 mT typically reported for modular reconfiguration under bulky 3D arrays \cite{Rogowski2023, Bhattacharjee2022}, demonstrating that present design is highly efficient. Operating below 13 mT reduces resistive heating and eliminates the need for active cooling systems \cite{Yang2020}, improving scalability and portability.

Self-assembly into chains occurred at 1.5 mT, with assembly time scaling approximately linearly with initial separation distance (\autoref{fig:ChTID}). This demonstrates controllable aggregation under weak fields, in contrast to the higher fields (10–40 mT) typically reported for modular reconfiguration using bulky 3D arrays \cite{Rogowski2023, Bhattacharjee2022}. Operating at $<13$ mT reduces resistive heating and eliminates the need for active cooling systems, improving scalability and portability  \cite{Yang2020}.

\subsubsection{Reconfiguration}
% Chain-to-gripper reconfigurations achieved 90\% success with an average completion time of $3.88 \pm 0.65$ s. This enabled the generation of a open-loop sequence for cargo manipulation, where two \textit{gripper} modules and one \textit{free} module, initially in their liquid state, self-assembled into a chain, reconfigured into a gripper for grasping the substance, and transported it across the workspace for cargo delivery at the goal location. Finally, the structure “walked” away after task completion (\autoref{fig:MMT}(f)). A demonstration of this reconfiguration is provided in the supplementary video. % (\nameref{app:Video1}) % Add video - submission after the 17th

Reconfiguration results were obtain from the data of 20 trials per transformation using fixed sequences with input currents adjusted for reconfiguration attempts. Chains were reliably reconfigured into grippers at 1.5–2.5 mT. Chain-to-gripper transformations achieved a 90\% success rate with an average completion time of $3.88 \pm 0.65$ s. The asymmetric geometry of the gripper required less torque for actuation and exhibited more reliable walking compared to the square configuration. This robustness makes the gripper particularly suitable for manipulation tasks.

The demonstrated gripper configuration enabled an open-loop cargo manipulation sequence. Two \textit{gripper} modules and one \textit{free} module, initially unbound, self-assembled into a chain, reconfigured into a gripper, grasped a target object, and transported it across the workspace for release at the goal location. Finally, the structure “walks” away after task completion. % Add video

% Conversely, chain-to-square transformations succeeded in 65\% of trials with longer completion times (6.7 s). This difference suggests that assembly geometry strongly affects reconfiguration robustness. The gripper design is lighter and mechanically asymmetric compared to square, required less torque for actuation and exhibited more reliable walking, whereas the square demanded higher fields of 4 mT. These findings are consistent with reports that symmetric structures are harder to reconfigure and actuate reliably under open-loop control \cite{Lu2023}.

Chains were also reconfigured into symmetric square structures, though success rates were lower (65\%) with longer average completion times (6.7 s). Square assemblies required higher fields (up to 4 mT) and were less stable under open-loop control. These findings are consistent with prior reports that symmetric structures are harder to reconfigure and actuate reliably under open-loop actuation \cite{Lu2023}.

\subsubsection{Disassembly}
Modules were disassembled at $\approx 12.6$ mT. This transition highlights the ability of the platform to toggle between strongly bound and freely moving states, enabling programmable assembly–disassembly cycles. Such reversibility is advantageous for modular microrobots in dynamic environments, where modules can detach for single-unit exploration or reconfigure for multitask completion.

% 5.3 Closed-Loop Maze Navigation
\subsection{Closed-Loop Maze Navigation}

The \textit{fixed} module successfully translated inside a maze using real-time vision feedback and A* path planning, representing the first demonstration of single-module closed-loop control of a magnetic millirobot in a two-dimensional electromagnetic system (\autoref{fig:Actuation}(c, d)). Previous studies of modular navigation involved rigid chains, larger 3D coil arrays, extended workspaces ($100 \times 100 ~\text{mm}^2$), and frequent pauses to manage coil heating of chain structures \cite{Rogowski2023}. By contrast, this work achieved single-module obstacle avoidance and goal reaching within an $8$ mm clearance environment using a compact 2D setup without cooling interruptions.

% 5.4 Limitations and Future Directions

\section{Discussions}
While the presented system demonstrates robust planar locomotion and reconfiguration, several limitations highlight opportunities for future research.
First, the current actuation setup does not directly support rolling-only locomotion. Although the hybrid “flip-and-walk” gait supports deterministic, multimodal motion in the plane, true rolling behavior has been enabled in prior magnetic microrobot designs using single-axis actuation \cite{Bhattacharjee2022}. Achieving rolling motion in this platform may be possible by applying field sequences that realign module dipoles along the vertical axis, or by further miniaturizing the modules (for example, toward $\sim1$ mm diameters), which could also facilitate navigation within tighter geometries.

Second, the present work does not investigate biological translation. In in vivo environments, complex fluid flows, tissue heterogeneity, immune responses, and mechanical constraints are absent from our simplified setup \cite{Lee2023}. Addressing these factors, including biocompatibility, immune evasion, and functional stability in physiological media, will be crucial for translating millirobotic systems to clinical contexts. 

Third, while maze navigation is demonstrated, it was evaluated qualitatively using a simple success criterion (reaching a target within $<$1 mm). Systematic metrics such as path-following error, correction frequency, and completion time were not recorded, limiting the ability to assess robustness in cluttered or dynamic environments. Incorporating well-tuned feedback controllers or model predictive control has been shown to improve navigation precision in magnetic microrobot tasks \cite{Jiang2022} and would enable more rigorous performance evaluation of single-module autonomous behaviors.

\section{Conclusions and Future Work}

This work presents a modular magnetic millirobotic platform capable of single-module locomotion, multimodal self-reconfiguration, and closed-loop maze navigation using a hardware-efficient two-dimensional electromagnetic setup operating below 13mT. By lowering actuation requirements relative to prior systems that rely on bulky three-dimensional coil infrastructures, the platform mitigates heating and scalability challenges while preserving functional versatility.
The system enables controlled, collision-free transformations into chain, square, and gripper configurations, demonstrating that modular architectures can support adaptive task execution in constrained environments without relying on boundary-assisted reconfiguration. Although the current prototype was evaluated within a limited workspace and under non-biocompatible conditions, these results indicate that low-field modular magnetic actuation is a promising direction for developing clinically relevant magnetic robotic systems.

Future work will focus on:
(i) quantitative benchmarking of closed-loop navigation performance, including success rates, tracking error, completion time, and robustness to disturbances;
(ii) miniaturization and incorporation of biocompatible materials and coatings to support in vivo operation; and
(iii) validation in physiologically realistic environments, including complex fluids and geometries representative of luminal or vascular anatomies.

%\addtolength{\textheight}{-12cm}

%%%%%%%%%%%%%%%%%%%%%%%%%%%%%%%%%%% APPENDIX %%%%%%%%%%%%%%%%%%%%%%%%%%%%%%%%%%%
\section*{APPENDIX}

%\subsection*{Supplementary Calculations}

The magnets are modeled as uniformly magnetized spheres with the following parameters \cite{NdFeB}:

\begin{itemize}
  \item Radius: \( r = {0.5}{mm} = {0.0005}{m} \)
  \item Magnetization: \( M = {986760}{A/m} \)
  \item Magnetic constant: \( \mu_0 = 4\pi \times 10^{-7} \, {T \cdot m/A} \)
  \item External field orientation: \( \theta = 90^\circ \Rightarrow \sin\theta = 1 \)
\end{itemize}

\subsubsection*{Volume}
\[
V = \frac{4}{3} \pi r^3 = \frac{4}{3} \pi (0.0005)^3 = {5.236 \times 10^{-10}}~{m^3}
\]

\subsubsection*{Magnetic Moment}
\[
m = M \cdot V = 986760 \cdot 5.236 \times 10^{-10} = {5.164 \times 10^{-4}}~{A \cdot m^2}
\]

\subsubsection*{External Fields} \label{app:ExtFields}
To estimate the minimum uniform field required to rotate one of the magnets, we compare the torque exerted by the external field with the torque generated by the magnetic interaction between the two magnets. A magnetic dipole \( \vec{m} \) in an external magnetic field \( \vec{B} \) experiences a torque given by \autoref{eq:Torque}. To initiate rotation, this torque must overcome the torque resulting from the magnetic field produced by the other magnet. Under the assumption that the magnets are small and the separation distance is relatively large compared to their size, each magnet can be modeled as a magnetic dipole. The magnetic field at a position \( \vec{r} \) from a dipole is given by the dipole field equation \cite{Seleznyova2016}:
\[
\vec{B}(\vec{r}) = \frac{\mu_0}{4\pi r^3} \left[ 3(\vec{m} \cdot \hat{r})\hat{r} - \vec{m} \right]
\]
where \( \hat{r} \) is the unit vector pointing from the dipole to the point of observation, and \( r = |\vec{r}| \).

The external field is applied perpendicular to the magnetic moment of the second magnet, and the magnets are separated by a center-to-center distance \( d \). For the point on the perpendicular bisector of the dipole (i.e., where the angle between \( \vec{m} \) and \( \hat{r} \) is \( 90^\circ \)), the dot product \( \vec{m} \cdot \hat{r} = 0 \), and the magnetic field magnitude simplifies to:
\[
B_{\text{dip}} = \left|\vec{B}(\vec{r})\right| = \frac{\mu_0}{4\pi} \cdot \frac{m}{d^3}
\]
Thus, to counteract the interaction and rotate the second magnet, the external field must satisfy:
\[
B_{\text{required}} \geq B_{\text{dip}} = \frac{\mu_0}{4\pi} \cdot \frac{m}{d^3}
\]

\paragraph*{1. Chain-to-Gripper Reconfiguration}

In the chain of two \textit{gripper} modules and one \textit{free} module, the largest separation occurs between magnets 2 and 3 (\autoref{fig:Motion}(c)):
d = 3.2~\text{mm} = 0.0032~\text{m}.

\begin{align}
B_{\text{required}} 
&= \frac{10^{-7} \cdot 5.164 \times 10^{-4}}{(0.0032)^3} \notag = \frac{5.164 \times 10^{-11}}{3.277 \times 10^{-8}} \notag \\
&= 0.00158~\text{T} = 1.58~\text{mT}.
\end{align}

Therefore, the minimum uniform field to induce gripper reconfiguration is \(\boxed{1.58~\text{mT}}\).

\paragraph*{2. Chain-to-Square Reconfiguration}

For a four-module chain reconfiguring into a square, the shortest separation between intermediate magnets (modules 2 and 3) is
d = 3.0~\text{mm} = 0.0030~\text{m}.

\begin{align}
B_{\text{required}} 
&= \frac{10^{-7} \cdot 5.164 \times 10^{-4}}{(0.0030)^3} \notag = \frac{5.164 \times 10^{-11}}{2.7 \times 10^{-8}} \notag \\
&= 0.00191~\text{T} = 1.91~\text{mT}.
\end{align}
Thus, reconfiguration into a square requires at least \(\boxed{1.91~\text{mT}}\).

\paragraph*{3. Disassembly}
For two modules forming the first bond in a chain, the center-to-center distance is d = 1.6~\text{mm} = 0.0016~\text{m}.
\begin{align}
B_{\text{required}} 
&= \frac{10^{-7} \cdot 5.164 \times 10^{-4}}{(0.0016)^3} \notag = \frac{5.164 \times 10^{-11}}{4.096 \times 10^{-9}} \notag \\
&= 0.0126~\text{T} = 12.6~\text{mT}.
\end{align}
Thus, a uniform field of at least \(\boxed{12.6~\text{mT}}\) is required to separate any structure.

% VIDEOS: Cannot submit until the 17-22nd, but these are the links:

% Cargo transport and release using a reconfigurable chain-to-gripper transformation: https://imperiallondon-my.sharepoint.com/:v:/g/personal/eg424_ic_ac_uk/EUvj8CuqnD9HjHyJ1dHeCZ0B9c8SSPvJl8Np4TacJyB0qQ?e=TGbpfi

% Single-Module Closed-Loop Maze Navigation: https://imperiallondon-my.sharepoint.com/:v:/g/personal/eg424_ic_ac_uk/EYphIJ8pB-RCqF1yGrmtsS0B4I4vNo-KhSvezYJ8ZDwYjQ?e=peaS9L

%%%%%%%%%%%%%%%%%%%%%%%%%%%%%%%% ACKNOWLEDGEMENT %%%%%%%%%%%%%%%%%%%%%%%%%%%%%%%
 \section*{ACKNOWLEDGMENT}

The authors would like to express sincere gratitude to Dr. Fabio Tatti, Mr. Niraj Kanabar, and Mr. QingZheng Cong for their advice on manufacturing and electrical setup.

%%%%%%%%%%%%%%%%%%%%%%%%%%%%%%%%%% REFERENCES %%%%%%%%%%%%%%%%%%%%%%%%%%%%%%%%%%
\bibliographystyle{IEEEtran}
\bibliography{refs}

\end{document}